\documentclass[letterpaper, 10 pt, conference]{ieeeconf}  

\IEEEoverridecommandlockouts                              
\usepackage[utf8]{inputenc}

\usepackage{subcaption}
\usepackage{amsmath}
\usepackage{amsfonts}
\usepackage{amssymb}
\usepackage{color}
\usepackage{mathptmx}       
\usepackage{helvet}         
\usepackage{courier}        
\usepackage{type1cm}        
\usepackage{makeidx}         
\usepackage{graphicx}        
\graphicspath{{Figures/}}
\usepackage{multicol}        
\usepackage[bottom]{footmisc}
\usepackage{paralist}        
\usepackage[normalem]{ulem}  

\usepackage[]{capt-of}
\usepackage{tikz}
\usetikzlibrary{arrows,backgrounds,calc,chains,external,fit,patterns,positioning,scopes,shapes.geometric,through}
\usepackage{pgfplots}
\pgfplotsset{compat=1.8}
\usepgfplotslibrary{colormaps,external,groupplots,statistics}
\usepackage[ruled]{algorithm2e}
\usepackage{booktabs}
\newlength\mylen
\newcommand\myinput[1]{%
	\settowidth\mylen{\KwIn{}}%
	\setlength\hangindent{\mylen}%
	\hspace*{\mylen}#1\\}

\DeclareMathOperator*{\argmin}{arg\,min}

\makeindex             

\begin{document}

\title{Functional Path Optimisation for Exploration in Continuous Occupancy Maps}
\author{Gilad Francis, Lionel Ott and Fabio Ramos{$^*$}
\thanks{$^*$ Gilad Francis, Lionel Ott and Fabio Ramos are with The School of Information Technologies, University of Sydney, Australia
        {\tt\small gilad.francis@sydney.edu.au}}%
}
%
\maketitle
\thispagestyle{empty}
\pagestyle{empty}

\begin{abstract}
Autonomous exploration is a complex task where the robot moves through an unknown environment with the goal of mapping it. The desired output of such a process is a sequence of paths that efficiently and safely minimise the uncertainty of the resulting map. However, optimising over the entire space of possible paths is computationally intractable. Therefore, most exploration methods relax the general problem by optimising a simpler one, for example finding the single next best view. 
In this work, we formulate exploration as a variational problem which allows us to directly optimise in the space of trajectories using functional gradient methods, searching for the \textit{Next Best Path} (NBP). We take advantage of the recently introduced Hilbert maps to devise an information-based functional that can be computed in closed-form. The resulting trajectories are continuous and maximise safety as well as mutual information. In experiments we verify the ability of the proposed method to find smooth and safe paths and compare these results with other exploration methods.
\end{abstract}

\section{Introduction}
\label{sec:intro}

The objective of autonomous exploration is to produce a consistent representation of the environment. It involves complex decision-making, selecting the trajectories a robot should take in order to minimise the overall uncertainty in the model. Essentially, exploration is a path optimisation procedure to find trajectories that efficiently learn the environment. The difficulty lies in the dimensionality and shape of the search space which prohibits a closed-form solution to the general exploration problem, making autonomous exploration an active field of research. The plethora of exploration methods in the literature offer different strategies for relaxing this intractable problem, most commonly by reducing the search space dimensionality, for example by discretising the path.

In this work, we formalise exploration as a variational problem. We present a novel approach based on \textit{functional gradient descent} (FGD) to efficiently optimise exploratory paths over continuous occupancy maps. We use stochastic FGD to overcome the limitations of standard FGD methods in order to ensure convergence. This process enables optimisation over the entire path, resulting in continuous smooth paths that maximise the overall map quality while keeping the robot safe from collisions. Our contributions are:
\begin{enumerate}
\item A \textit{Next Best Path} method. An information-driven variational framework for safe autonomous exploration in continuous occupancy maps. Path optimisation is performed directly in the space of trajectories using a combined objective; which considers safety, efficiency and information. The method is invariant to the choice of path representation as it uses stochastic functional gradient descent to optimise the objective along the entire path.
\item Developing a \textit{mutual information} (MI) variational objective for continuous occupancy maps. This replaces the common and expensive approach of computing MI explicitly over the entire map for each evaluated path. Instead, our method modifies the path using MI functional gradients without the need to compute MI explicitly. These gradients are obtained from local perturbations on the map model which are derived in closed-form.
\end{enumerate}

The remainder of this paper is organised as follows.
Literature on autonomous exploration is surveyed in Section \ref{sec:literature}. The basic building blocks, such as Hilbert maps and FGD are reviewed in Section \ref{sec:preliminaries}. Section \ref{sec:methods} describes in detail the functional exploration algorithm. Experimental results and analysis are presented in Section \ref{sec:results}. Finally, Section \ref{sec:conclusions} draws conclusions on the proposed method.

\section{Related Work}
\label{sec:literature}

The goal of autonomous exploration is to produce a consistent environment model by minimising any uncertainties. In a mapping context, exploration is the process of producing high-fidelity maps \cite{Stachniss2009}. This is a complex problem mainly due the dimensionality of the solution space. Most exploration methods use occupancy grid maps in their planning \cite{Thrun2005a}, rather than continuous occupancy maps. Regardless of the type of occupancy map, exploration methods take one of two forms; frontier-driven or information-theoretic.  Juli\'a et al. \cite{Julia2012} provide a quantitative comparison between these exploration methods.

Frontier-based exploration methods drive the robot toward the borders of the known space \cite{Yamauchi1997}. In a grid map, frontiers are clusters of free cells neighbouring unknown cells. Once frontiers are identified, a separate path planner finds a safe path toward a selected frontier. Various utility functions can be used when choosing the most desirable frontier. The simplest form considers only the travelling costs \cite{Yamauchi1997}. Gonz\'alez-Ba\~nos and Latombe \cite{Gonzalez-Banos2002} choose a goal point based on a score of expected coverage penalised by the travelling costs. A generalised approach for goal point selection given several criteria were suggested by \cite{Holz2011}. 

Information-theoretic exploration methods optimise a utility function associated with the uncertainty of the map. Early work optimised the selection of a discrete goal point rather than optimising an entire path. Elfes \cite{Elfes1996} suggested MI as an information metric for exploration, while \cite{Whaite1997} proposed a \textit{next best view} (NBV) approach using the entire map entropy. Vallv\'e and Andrade-Cetto  \cite{Vallve2015} use a potential field computed over the entire configuration space to find exploration candidates. However, this method assumes discrete steps, disregarding the reduction in entropy between consecutive robot poses. 
Charrow et al. \cite{Charrow-RSS-15} combined frontier and information-based methods. While they optimised the information heuristic over a continuous control input space, in effect the path consisted of a fixed number of time steps. Lauri and Ritala \cite{Lauri2015} formulate the exploration problem as \textit{partially observable Markov decision process} (POMDP) and used sample-based approach to solve the POMDP. Similarly to the work of \cite{Charrow-RSS-15}, the action space is continuous, however, the path consists of a finite set of time steps, which is in contrast to the proposed method where optimisation is performed in the space of trajectories.

Only a handful of algorithms tackle exploration in continuous occupancy maps. Yang et al. \cite{Yang2013} employed \textit{rapidly-exploring random tree} (RRT) to sample a set of feasible path candidates for a \textit{Gaussian Processes} (GPs) maps. An adaptation of the frontier method for continuous occupancy maps was introduced by \cite{Jadidi2014}, where a discretised frontier map was built from the continuous map. More recently, Jadidi et al. \cite{Jadidi2015} employed MI to rank these frontiers. \textit{Bayesian optimisation} (BO) has also been used for exploration. Marchant and Ramos \cite{Marchant2014} utilised BO to optimise the selection of continuous informative paths over a continuous environmental model. Francis et al. \cite{Francis2017a} used constrained BO for safe exploration to learn an MI objective. While BO optimises over continuous paths, in practice it uses only a limited path parameterisation such as quadratic or cubic splines. 

In summary, the exploration method proposed in this paper uses an information-based utility to optimise path selection. Employing calculus of variations, the optimisation procedure is invariant to the path representation. This is a major difference from existing methods that typically depend on a finite path parametrisation, such as finite sets of time steps or waypoints, or employ simple representations such as quadratic or cubic splines. Instead, our method can utilise a highly expressive path representation; such as non-parametric \cite{Francis2017} or approximate kernel paths \cite{Francis2017b}.
As our method uses continuous occupancy maps, the MI utility can be derived directly from the map model in closed form, which simplifies computations.

\section{Preliminaries}
\label{sec:preliminaries}

In this section, we review the basic building blocks of the functional exploration method;  Hilbert maps and functional gradient path planning. In section \ref{sec:methods}, we adapt  these building blocks to support an information-driven safe exploration algorithm.

\subsection{Hilbert Maps}
\label{subsec:hilbert_maps}

A Hilbert map \cite{ramos2015hilbert} is a continuous discriminative model that predicts occupancy based on sensors observations. Unlike grid maps \cite{Elfes1989} that discretise space into a set of independent cells, Hilbert maps maintain neighbourhood information. Other continuous mapping methods based on \textit{Gaussian Processes} (GPs) \cite{ Jadidi2014, OCallaghan2012} hold similar properties. However, these methods are limited by the scalability of the associated GPs. Hilbert maps, on the other hand, use an approximate kernel logistic regression model, which renders updating and querying the map independent of the dataset size. In addition, \textit{stochastic gradient descent} (SGD) is used to enable real-time performance.

Formally, a Hilbert map is a \textit{logistic regression} (LR) classifier model that predicts occupancy anywhere in the map. It uses nonlinear projection into a \textit{reproducing kernel Hilbert space} (RKHS) in order to represent a real environment. Given a set of weights $\mathbf{w}$, the predictive occupancy posterior transforms into:
\begin{equation}\label{HMAP:Prediction}
	p(y = +1|\mathbf{x},\mathbf{}{w}) = 
	\frac{1}{1+\exp(-\mathbf{w}^T \hat{\Phi}(\mathbf{}{x}))} 
	=  \sigma(\mathbf{w}^T \hat{\Phi}(\mathbf{x})).
\end{equation}
In an occupancy map context $\mathbf{x} \in \mathbb{R}^D$ represents either a $2D$ or $3D$ location, while $y \in \{-1,+1\}$ denotes two possible binary outputs, unoccupied and occupied. We define $\sigma$ as the logistic sigmoid function. $\hat{\Phi}(\cdot)$ is a set of features that, in expectation, approximate the inner product defined by the kernel function $k(\cdot, \cdot)$. This approximation enables fast training of a map model, suitable for use in a robotics setting. There are several methods to approximate the kernel matrix \cite{ramos2015hilbert}. Given the desired approximation, the weights vector  $\mathbf{w}$ can be then trained by minimising the regularised \textit{negative log-likelihood} (NLL) as commonly performed in logistic regression methods \cite{ramos2015hilbert}.

\subsection{Functional Gradient Descent}
\label{subsec:FGD}
\textit{Functional gradient descent} (FGD) is a variational framework to optimise nonlinear models. It has been successfully applied to motion planning problem in recent years with the main objective of producing safe, collision-free paths. It was recently suggested as an alternative approach to sampling-based methods for path planning using occupancy maps \cite{Francis2017}. In this section, the general method is discussed, before the extension for autonomous exploration is described in section \ref{sec:methods}.

We first introduce notation. A path, $\xi: [0,1] \rightarrow {C}\in\mathbb{R}^D$, is a function that maps a time-like parameter $t \in [0,1]$ into configuration space ${C}$. The objective functional $U(\xi): \Xi \rightarrow \mathbb{R}$ returns a real number for each path $\xi \in \Xi$, corresponding to a cost or loss associated with $\xi$. $U(\xi)$ captures path properties such as smoothness and safety. The goal of the optimisation process is to find a path that minimises the overall costs:
\begin{equation}\label{eq:general_object}
\xi_{optimal}= \argmin_{\xi} U(\xi)
\end{equation}

Finding the optimal path is performed by following the functional gradient of the objective. This is an iterative process where the functional gradient update rule is derived from a linear approximation of the cost functional around the current trajectory, $\xi_n$:
\begin{equation}
	U(\xi) \approx U(\xi_n)+\nabla_\xi U(\xi_n)(\xi-\xi_n) .
\end{equation}
We enforce small updates by adding a regularisation term based on the norm of the update:
\begin{equation}\label{eq:FGMP_optimisaton}
	\xi_{n+1}= \argmin_\xi \quad U(\xi_n) + (\xi-\xi_n)^T\nabla_\xi U(\xi_n) + \frac{1}{2\eta_n}\|\xi-\xi_n\|_\Lambda^2.
\end{equation}
The regularisation term $\|\xi-\xi_n\|_\Lambda^2 = (\xi-\xi_n)^T\Lambda(\xi-\xi_n)$ is the squared norm with respect to a metric tensor $\Lambda$ and $\eta_n$ is a user-defined learning rate. By differentiating the right hand side of (\ref{eq:FGMP_optimisaton}) with respect to $\xi$, we obtain the iterative update rule:  
\begin{equation}\label{eq:FGMP_update_rule}
	\xi_{n+1}(\cdot)= \xi_n(\cdot) - \eta_n \Lambda^{-1}\nabla_\xi U(\xi_n).
\end{equation}
We note that (\ref{eq:FGMP_update_rule}) forms a general update rule, regardless of the choice of the objective function or the path representation. The only requirements are that $\Lambda$ is invertible and the gradient $\nabla_\xi{U}(\xi_n)$ exists. 

The general rule for computing the objective functional gradient $\nabla_\xi{U}$ stems from the calculus of variations. As a variational method, the objective functional must take the form of an integral or sum. Generally, the objective functional ${F}$ takes the form ${F}(\xi)=\int_{a}^{b} v(t,\xi,\xi') dt$, therefore the functional gradient can be computed using the Euler-Lagrange equation \cite{Zucker2013}:
\begin{equation}\label{eq:functional_grad}
	\nabla{F}_\xi(\xi)=\frac{\partial v}{\partial\xi} - \frac{d}{dt}\frac{\partial v}{\partial\xi'}.
\end{equation}
These gradients are then used to compute the iterative update rule of (\ref{eq:FGMP_update_rule}). Constraints are incorporated using KKT conditions, similar to \cite{Marinho2016}, and are an explicit part of the path representation.

\section{Functional Exploration}
\label{sec:methods}
The following section introduces our proposed functional exploration method. The use of functional gradient descent on an information-based objective results in trajectories which are highly expressive and safe while optimising the amount of information gained along the path.

\subsection{Notation}
\label{subsec:exploration_notation}
We first introduce the notation used throughout the following sections. The workspace of the robot ${W} \in \mathbb{R}^3$ defines the space where obstacles lie and the map is queried. In addition, to account for the robot's finite size or its pose uncertainty, a set of body points, ${B} \in \mathbb{R}^3$ are defined. As the trajectory $\xi$ lies in configuration space ${C}$, we set a forward kinematic transform $g$ that maps a robot configuration, $\xi(t)$ and body point $b \in {B}$ to a point in the workspace $g: {C} \times {B} \rightarrow {W}$. To simplify notation, we define $x_t=g(\xi(t),b)$ as the workspace location for the pair $(t,b)$. We assume that the robot is equipped with a sensor, such as laser range finder, with a maximum range $R_{max}$ and an angular field of view $\Omega$.

For a given function $\xi$, a functional returns a single value ${U}: \mathbb{R}^D\rightarrow \mathbb{R}$. Functionals are usually represented by an integral. However, whenever we compute the safety or MI functionals, the cost is computed over ${W}$. As a result, the functional must be approximated by a \textit{reduce} operator, e.g., average or  maximum, that aggregates the cost along $\xi(\cdot)$. Given a workspace cost function $c\Big(g\big(\xi(t),b\big)\Big): \mathbb{R}^3 \rightarrow \mathbb{R}$, we can approximate the functional by a sum over a finite set $\mathcal{T}(\xi) = \{t,b\}_i$ of time and body points:
\begin{multline}\label{eq:approximated_U}
{U}_{obs,MI}(\xi) \approx \sum_{(t,b)\in\mathcal{T}(\xi)} c_{obs,MI}\Big(g\big(\xi(t),b\big)\Big) \equiv \\ \sum_{(t,b)\in\mathcal{T}(\xi)} c_{obs,MI}(x_t).
\end{multline}
In addition, we note the difference between gradient operators. We define $\nabla$ as a gradient with respect to $\xi$, while $\nabla_x$ is the workspace gradient.

\subsection{Exploration Functional Objective}
\label{subsec:exploration_objective}

The goal of autonomous exploration is to safely reduce the uncertainty of the environment model. Some exploration methods compute a finite set of go-to points that locally maximise information gain. Other methods optimise an information-based objective over the entire path, however, these are highly dependent on the path parameterisation. Formulating exploration as a variational problem and solving it using functional gradient descent, provides a general optimisation framework which is invariant to the choice of path parameterisation and can even take a non-parametric form as shown in \cite{Francis2017}.

\begin{figure}[tb] 
    \centering
	
    \includegraphics[width=0.48\textwidth]{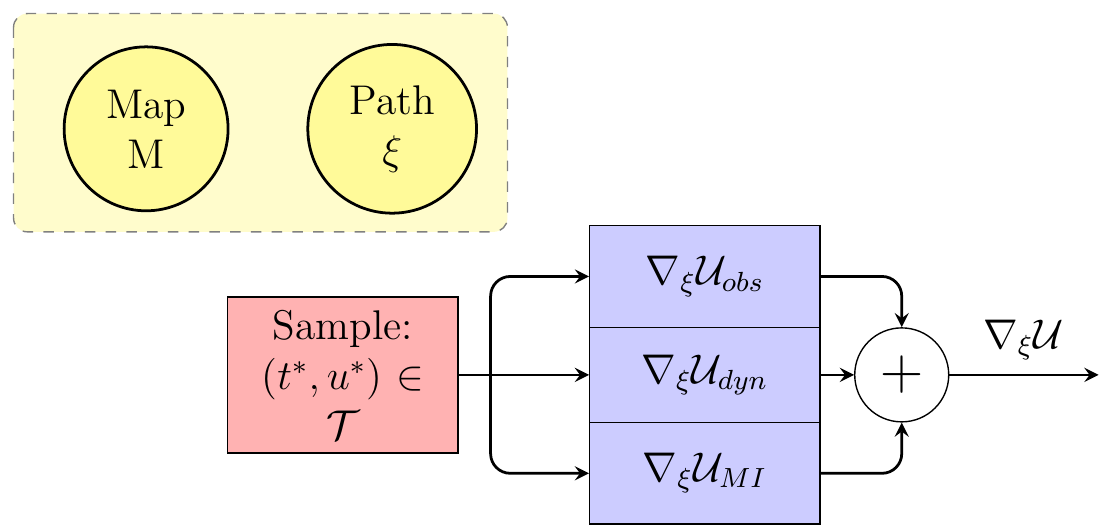}
    \caption{Given a sample $(t, b)$, the functional $U$ is the weighted sum of the various objectives; obstacle, dynamic and MI. $M$ and $\xi$ are global variables used to compute these objectives.}
    \label{fig:FGD}
    
\end{figure}

The approach taken in our work relies on maximising mutual information along the entire path while keeping the trajectory safe. This is attained by constructing an objective functional ${U}$ which is a weighted sum of three components:
\begin{itemize}
\item ${U}_{obs}$ which maintains path safety by penalising proximity to obstacles,
\item ${U}_{dyn}$ which penalises based on the shape of the trajectory, keeping path smooth and short and,
\item ${U}_{MI}$ which rewards the mutual information gained along the path.
\end{itemize}
The overall objective takes a form of a weighted sum, as is also shown schematically in Figure \ref{fig:FGD};
 \begin{equation}\label{eq:FPMP_U}
{U}(\xi) = \beta_{obs} {U}_{obs}(\xi) + \beta_{dyn} {U}_{dyn}(\xi)+ \beta_{MI} {U}_{MI}(\xi).
\end{equation}
Here $\beta_{obs}$,$\beta_{dyn}$,$\beta_{MI}$ are user-defined coefficients.
In the following sections, we will introduce the different components of the functional objective; ${U}_{obs}(\xi)$, ${U}_{dyn}(\xi)$ and ${U}_{MI}(\xi)$. For each component we will derive its functional gradient assuming a Hilbert map as the environment model.

\subsubsection{Obstacle Functional ${U}_{obs}(\xi)$} \label{subsubsec:obs_fnctl}

Following our previous work \cite{Francis2017}, we define the workspace cost function $c_{obs}$ as the map occupancy of (\ref{HMAP:Prediction}), i.e. $c_{obs}(x_t) = p(y = +1|x_t)$. Given (\ref{eq:approximated_U}),  the obstacle functional can be approximated by 
${U}_{obs}(\xi) \approx \sum_{x_t\in\mathcal{T}(\xi)} c_{obs}(x_t)$. The functional gradient can then be computed as
\begin{equation}\label{eq:d_U_obs-final}
\nabla{U}_{obs}=\sum_{x_t \in \mathcal{T}(\xi)} \mathbf{J}(x_t) \nabla_x c_{obs}(x_t).
\end{equation}
Here, $\mathbf{J}(x_t)$ is the workspace Jacobian. Since the map model is continuous and at least twice differentiable \cite{ramos2015hilbert}, the spatial gradient of occupancy can be computed in closed-form from (\ref{HMAP:Prediction}) as:
\begin{multline}\label{eq:hmap_gradient}
\nabla_x c_{obs}(x_t) = \nabla_x p(y = +1|x_t) = \\ \sigma\Big(\mathbf{w}^T \hat{\Phi}(x_t)\Big)\Big(1- \sigma\big(\mathbf{w}^T \hat{\Phi}(x_t)\big)\Big)\mathbf{w}^T \nabla_x \hat{\Phi}(x_t).
\end{multline}

\subsubsection{Path Dynamics Functional ${U}_{dyn}(\xi)$} \label{subsubsec:dyn_fnctl}
${U}_{dyn}(\xi)$ penalises kinematic costs associated with $\xi$. The straightforward approach is to regularise on the trajectory length, which can be attained by optimising the integral over the squared velocity norm: ${U}_{dyn}(\xi) = \frac{1}{2}\int_{0}^{1} \left|\left|\frac{d}{dt}\xi(t) \right|\right|^2 dt$. Following the Euler-Lagrange equation (\ref{eq:functional_grad}), the functional gradient of ${U}_{dyn}$ is 
\begin{equation}\label{eq:d_U_reg-final}
	\nabla_\xi {U}_{dyn}(\xi(t))= -\frac{d^2}{dt^2}\xi(t).
\end{equation}

\subsubsection{Mutual Information Functional ${U}_{MI}(\xi)$} \label{subsubsec:MI_fnctl}
 \begin{figure*}[thpb]
	
	\centering
	
	\includegraphics[width=0.90\textwidth]{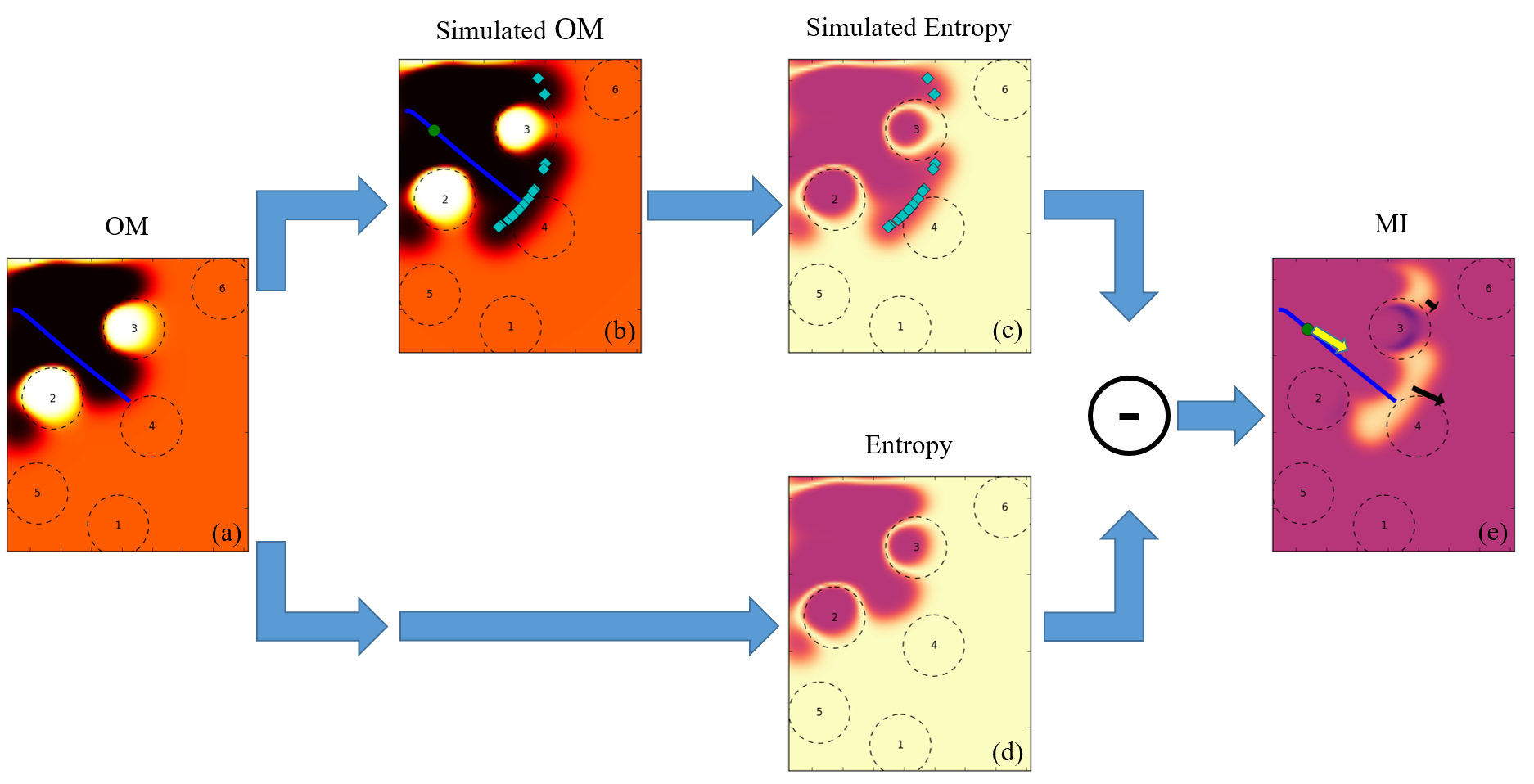}
	
	\caption{MI Functional gradient. The MI gradient is computed for a time sample $t$ given a continuous occupancy map $OM$ (a). The current path estimate $\xi$ is depicted in blue, while the expected pose at time $t$ is shown in green. The modified OM (b) is generated using hypothetical observations, shown as cyan diamonds, based on a robot's configuration at time $t$. Only observations at the edge of sensor range are considered for $t$. The entropy of the OM (c) and the modified OM (d) can be computed from the occupancy values (high entropy shown in white), which produces MI (e). The MI gradient is estimated by samples around observations (shown as black arrows). Accumulating gradient samples results in the overall MI gradient for $t$, shown by the yellow arrow on the robot. Note that the images of occupancy, entropy and MI are only given here for presentation purposes and full maps are never computed. The planner only accesses occupancy, entropy and gradient values through stochastic samples.}
	\label{fig:MI_calc}
\end{figure*}

\textit{Mutual information} (MI) is used in many autonomous exploration algorithms as an information-based objective function \cite{Julian2014}. In this context, MI is defined as the reduction in entropy conditioned on expected observations. Given an occupancy map $M$ and a set of expected observations $\hat{\mathbf{z}}$, we can define MI as: 
\begin{equation}\label{eq:MI}
	MI(M; \hat{\mathbf{z}}) = H(M) - H(M | \hat{\mathbf{z}})
\end{equation} 
where $H$ denotes Shannon's entropy. The main computational challenge of (\ref{eq:MI}) is resolving the expected observations $\hat{\mathbf{z}}$. These are emulated observations that are produced by ray casting based on the sensor model. Determining $\hat{\mathbf{z}}$ and MI over an entire path is computationally intensive, leading most exploration methods to solve a relaxed MI optimisation problem, by either discretising or parameterising the paths.    

The approach taken in this work uses the MI functional ${U}_{MI}$ and its gradient to maximise MI efficiently over the entire trajectory. To compute ${U}_{MI}$, an MI reward function $c_{MI}(x_t) \approx MI(M; \hat{\mathbf{z}} | x_t)$ is computed over the robot's workspace ${W},$ in a similar fashion to the computation of ${U}_{obs}$. However, computation of the conditional entropy $H(M | \hat{\mathbf{z}})$ entails changes to the Hilbert map model. 

In the following section, we will describe the stages involved in computing the MI functional gradient. The three stages executed when computing $\nabla {U}_{MI}$ are:
\begin{itemize}
\item Simulating expected observations
\item Creating a perturbed Hilbert map model $M | \hat{\mathbf{z}}$
\item Obtaining MI functional gradient $\nabla {U}_{MI}$
\end{itemize}
Fig. \ref{fig:MI_calc} shows the steps required for MI gradient computations at a given time $t$.

\noindent \textbf{ Simulating expected observations}

\begin{figure}[!t]
 \centering
\subcaptionbox{Standard Exploration\label{RKHS:subfig-1}}
{\includegraphics[width=0.27\textwidth]{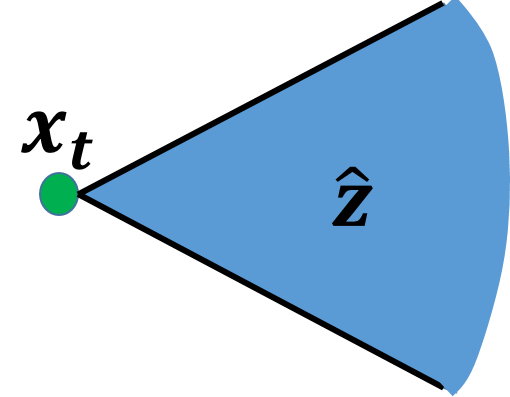}}
\subcaptionbox{Functional Exploration\label{RKHS:subfig-2}}
{\includegraphics[width=0.27\textwidth]{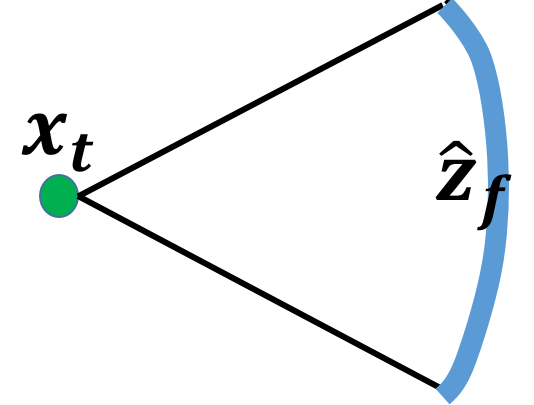}}
 \caption{Difference in MI calculation. $x_t$ is the robot's pose for which MI is computed, the black lines depict the sensor field of view and the blue area is the region where MI is computed. \protect\subref{RKHS:subfig-1}  MI is computed over the entire field of view of sensor, producing $\hat{\mathbf{z}}$. \protect\subref{RKHS:subfig-2} MI is computed only at the sensor's range limits, producing $\hat{\mathbf{z}}_f$}
 	\label{fig:FOV}
\end{figure}

Similarly to the obstacle functional, the MI functional is approximated by a sum over a finite set of points $x_t$; ${U}_{MI}(\xi) \approx \sum_{x_t \in\mathcal{T}(\xi)} c_{MI}(x_t)$. $c_{MI}$ is chosen in a way that will estimate the infinitesimal change in MI at $x_t$.
To compute $c_{MI}$ we emulate observations by ray casting, as done in other information-driven exploration techniques \cite{Elfes1996,Jadidi2015}. However, while other methods evaluate the MI reward over the entire field of view of the sensor, our method is only interested in the expected observations at the sensor's limits (maximum range). The rationale behind this approach is that new information about the environment will be mainly obtained at the sensor's sensing limits $R_{max}$. Fig. \ref{fig:FOV} illustrates the difference between the two approach to compute MI.

The output of the ray casting process is a set of unoccupied expected observations $\hat{\mathbf{z}}_{f}$ for any time and body point $x_t$. $\hat{\mathbf{z}}_{f}$ differs from the expected observations of (\ref{eq:MI}), as we are only interested in unoccupied (no obstacle) observation at the sensor maximum range. Fig. \ref{fig:MI_calc}b depicts $\hat{\mathbf{z}}_{f}$ as cyan diamonds.
 
\begin{algorithm}[tb]
	\caption{Stochastic Functional Exploration}
	\label{algo:main}
	\DontPrintSemicolon
	\KwIn{${M}$: Occupancy Map.}
	\myinput{$\xi(0)$: Start state.}
	\myinput{$P_{safe}$: No obstacle threshold.}
    \myinput{Optional: boundary conditions $\xi_b$, initial guess $\xi_o$.}

	\KwOut{$\mathbf{w}_{optimal}$}

	$n=0$\\
	\While{solution not converged }
	{
		//Stochastic sampling:	\\
		$S \sim U[0,1]$ Draw mini-batch \\
		\ForEach{$x_t \in S$}
		{
        $P_{occ} \leftarrow M(x_t))$ Eq. (\ref{HMAP:Prediction})\\
			\If{$P_{occ} \leq P_{Safe}$}
            {
            $\hat{\mathbf{z}}_{f}(x_t) \leftarrow$  simulated observations\\
            $\nabla{U}_{obs} \leftarrow  \frac{\partial}{\partial \xi}x_t \big(\sigma\Big(\mathbf{w}^T \hat{\Phi}(x_t)\Big)\Big(1- \sigma\big(\mathbf{w}^T \hat{\Phi}(x_t)\big)\Big)\mathbf{w}^T \nabla_x \hat{\Phi}(x_t)\big)$  Eq.(\ref{eq:d_U_obs-final}) \\
            $\nabla{U}_{dyn} \leftarrow -\frac{d^2}{dt^2}\xi(t)$ Eq.(\ref{eq:d_U_reg-final})\\
            $\nabla{U}_{MI} \leftarrow \sum_{m \in m_{max}(x_t)} \frac{\partial}{\partial \xi}x_t \nabla_x MI\big(M(m); \hat{\mathbf{z}}_{f}(x_t)\big)$ Eq.(\ref{eq:d_U_MI-final})\\
            
            $\nabla{U} \leftarrow$ Combine using Eq.(\ref{eq:FPMP_U})\\
            
            $\mathbf{w}_{n+1} \leftarrow \mathbf{w}^\xi_n - \eta_n {\Lambda^{-1} \hat{\Upsilon}(t_i)^T} \hat{\Upsilon}(t_i)\nabla_\xi{U}(\xi_n)(t_i)$ update rule Eq. \ref{eq:update_rule_weights}\\
            }
        }
		$n=n+1$	
	}
\end{algorithm} 

\noindent \textbf{Creating a perturbed Hilbert map model $M | \hat{\mathbf{z}}$}

In this step, we generate $M |\hat{\mathbf{z}}_{f}$, the modified Hilbert map conditioned on the expected observations. The straightforward approach is to train a new map model based on $\hat{\mathbf{z}}_{f}$. This approach is commonly used in exploration methods for occupancy grid maps. However the computational costs of such an approach are high, as new maps must be generated along the entire trajectory during optimisation. Instead, we propose the use of a perturbed Hilbert map model.

The perturbed Hilbert map model modifies the predictive map model $\sigma\big(\mathbf{w}^T \hat{\Phi}(\mathbf{x})\big)$ (\ref{HMAP:Prediction}) with a perturbed model $\sigma\big(\Psi(\mathbf{x},\hat{\mathbf{z}}_{f})\big)$. This model uses the expected observations as a dataset $\hat{\mathbf{z}}_{f}= \big\{x_i,r_i\big\}_{i=1}^N$, where $x_i \in \mathbb{R}^D$ is a point in 2D or 3D space and $r_i$ is the log odds of the desired predictive occupancy posterior at $x_i$, to fit a \textit{Gaussian process} (GP):
\begin{equation}
\Psi(\mathbf{x},\hat{\mathbf{z}}_{f}) \sim GP\big(\mathbf{w}^T \hat{\Phi}(\mathbf{x}), k(x_i, x_i)\big).
\end{equation} We note that $\mathbf{w}^T \hat{\Phi}(\mathbf{x})$ of the current occupancy map is the mean function of the GP. The kernel function $k$ is the same function approximated by the Hilbert map features $\hat{\Phi}(\cdot)$. The predictive probability of the perturbed map $\hat{p}$ is given by:
\begin{equation}\label{eq:HMAP:perturbed_map}
\hat{p}(\mathbf{x},\hat{\mathbf{z}}_{f}) = \sigma\big(\Psi(\mathbf{x},\hat{\mathbf{z}}_{f})\big)
\end{equation}
Fig. \ref{fig:MI_calc}b depicts the resulting occupancy map following the embedding of the expected observations in an existing map. 

The computational cost of the perturbed Hilbert map is cubic in the number of expected observations $|\hat{\mathbf{z}}_{f}|$. As we are only modifying the map in a small region, a small set of observations is required to generate perturbation, keeping the computation load low.

\noindent \textbf{Obtaining MI functional gradient $\nabla {U}_{MI}$}

The workspace MI cost function for $x_t$ is defined as the MI summed over the entire map, i.e. 
$c_{MI}(x_t) = \int_{M} MI\big(M(m); \hat{\mathbf{z}}_{f}(x_t)\big)dm$. However, as we are only interested in the change at the limits of the sensor range, $c_{MI}$ can be replaced by; $c_{MI}(x_t) = \int_{m \in m_{max}(x_t)} MI\big(M(m); \hat{\mathbf{z}}_{f}(x_t)\big)dm$. Where $m_{max}(x_t)$ denotes workspace locations which lie on the arc given by the maximum sensing range $R_{max}$ and the sensor's field of view $\Omega$, as shown in Fig. \ref{fig:FOV} . $c_{MI}$ can be approximated with a sum using either a deterministic or a Monte-Carlo schedule:
\begin{equation}\label{eq:c_MI_sum}
c_{MI}(x_t) \approx \sum_{m \in m_{max}(x_t)} MI\big(M(m); \hat{\mathbf{z}}_{f}(x_t)\big).
\end{equation}
Given the approximations (\ref{eq:approximated_U}) and (\ref{eq:c_MI_sum}), the MI functional can be represented as 
\begin{equation}\label{eq:U_MI_final}
{U}_{MI}(\xi) \approx \sum_{x_t\in\mathcal{T}(\xi)} \sum_{m \in m_{max}(x_t)} MI\big(M(m); \hat{\mathbf{z}}_{f}(x_t)\big).
\end{equation}

The MI functional gradient follows the same form as (\ref{eq:d_U_obs-final}):

\begin{equation}\label{eq:d_U_MI-final}
\nabla{U}_{MI}(x_t)=\sum_{x_t\in\mathcal{T}(\xi)} \sum_{m \in m_{max}(x_t)} \mathbf{J}(x_t) \nabla_x MI\big(M(m); \hat{\mathbf{z}}_{f}(x_t)\big).
\end{equation}

The spatial gradient of MI, $\nabla_x MI\big(M(m)$, can be expressed in closed-form using the Hilbert maps' continuous model. Using the MI definition from (\ref{eq:MI}), the gradient is defined as $\nabla_x MI(M; \hat{\mathbf{z}}) = \nabla_x H(M) - \nabla_x H(M | \hat{\mathbf{z}})$, where $\nabla_x H$ is the spatial gradient of the entropy.
Using the chain rule we rewrite the spatial gradient of $H$ around a query point $x$ as:
\begin{equation}\label{eq:H_grad}
\nabla_x H(x) = \frac{dH}{dp}\nabla_x p(x),
\end{equation}
where $p$ is the probability of occupancy at $x$ given by (\ref{HMAP:Prediction}). As $p$ is a Bernoulli random variable, $\frac{dH}{dp}$ is simply; $\frac{dH}{dp} = log_{_2}\frac{1-p}{p}$.
The occupancy gradient $\nabla_x p(x)$ depends on the occupancy map used. For the unperturbed map, $\nabla_x p(x)$ is given by (\ref{eq:hmap_gradient}). The spatial gradient of the perturbed map (\ref{eq:HMAP:perturbed_map}) can be computed similarly to the unperturbed map;
\begin{equation}\label{eq:HMAP:perturbed_map_grad}
\begin{gathered}
\nabla_x \hat{p}(\mathbf{x},\hat{\mathbf{z}}_{f}) = \nabla_x \sigma\big(\Psi(\mathbf{x},\hat{\mathbf{z}}_{f})\big) \\
= \sigma\big(\Psi(\mathbf{x},\hat{\mathbf{z}}_{f})\big) \Big(1-\sigma\big(\Psi(\mathbf{x},\hat{\mathbf{z}}_{f})\big) \Big) \nabla_x \Psi(\mathbf{x},\hat{\mathbf{z}}_{f}) \\
\end{gathered}
\end{equation}
As $\Psi$ is a GP, its gradient can be computed in closed form.

Fig. \ref{fig:MI_calc}e shows the computation of MI gradient by sampling along the arc defined by the sensor range. Each sample produces an MI gradient, schematically shown by the black arrows. The overall MI gradient, which pushes optimisation toward exploratory trajectories is computed from the sum of all samples and is shown in Fig. \ref{fig:MI_calc}e as a yellow arrow.

\begin{figure*}[tb]
	
	\centering
	
	\includegraphics[width=0.9\textwidth]{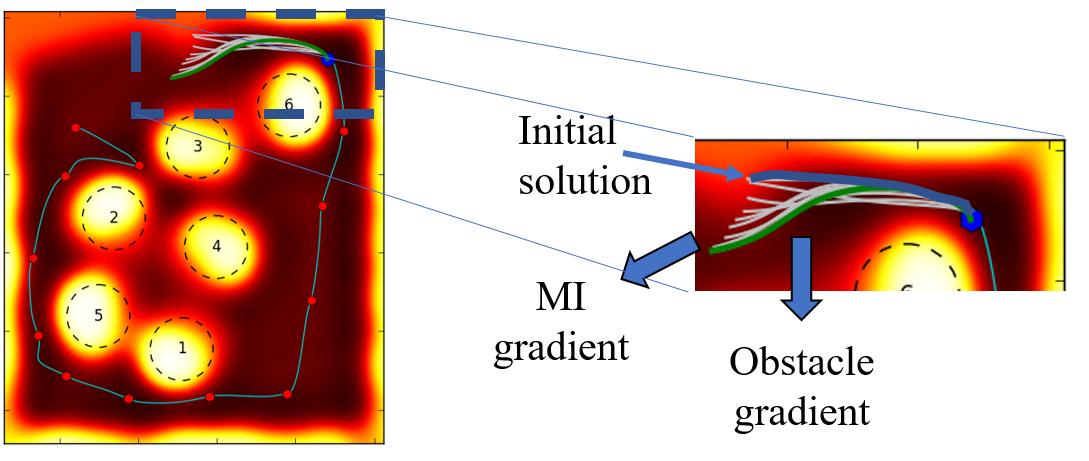}
	
	\caption{Functional planning iteration. Functional gradients deform the initial solution in blue. The obstacles gradient repels path from obstacles and unknown space. The MI gradient pulls path toward unexplored space. The resulting path, in green, is the optimised solution.}
	\label{fig:iteration}
\end{figure*}

\subsection{Functional Exploration Algorithm}

In this section, we describe the functional exploration algorithm, which aims to find a safe path that maximises MI over its entire course. 

Functional exploration is a general optimisation framework, meaning it is invariant to the choice of path representation. Functional optimisation methods have used waypoint parameterisation \cite{Zucker2013}, Gaussian process representation \cite{mukadam2016gaussian,Dong2016}, or defined trajectories over RKHS \cite{Marinho2016}. However, in all methods the objective is sampled via a deterministic schedule. Such approaches have proved unsatisfactory for planning using occupancy maps \cite{Francis2017}. To ensure convergence to a safe solution, the path is stochastically sampled in the entire $t \in [0,1]$ domain and any uninformative gradient updates, such as those coming from unsafe parts of the map, are rejected. 

The path representation used in this work is based on kernel matrix approximations \cite{Francis2017b}. The path is essentially a weighted sum of nonlinear features, similar to a regression problem. This path model is highly expressive, yet concise. Most importantly, the path can be optimised by SGD \cite{Robbins1951}. Given a set of weights $\mathbf{w}^\xi$, the path is defined as:
\begin{equation}\label{eq:approx_krnl_path}
	\xi(t) = (\mathbf{w}^\xi)^T\hat{\Upsilon}(t) + \xi_{o}(t) + \xi_{b}(t).
\end{equation}
$\xi_o$ is an initial path, which can be randomly chosen or can be computed by a simple and fast planner. $\xi_b$ is a term used to adjust boundary conditions, such as the start and goal pose. $\hat{\Upsilon}(\cdot)$ are nonlinear features that approximates the inner product defined by a kernel function $k_p$ by the following dot product \cite{Francis2017b};
\begin{equation} \label{eq:kernel_approx}
	k_p(t,t') = \langle\Upsilon(t),\Upsilon(t')\rangle \approx \hat{\Upsilon}(t)^T\hat{\Upsilon}(t').
\end{equation}
The kernel $k_p(t,t')$ maintains the correlation between various time points. 

Using (\ref{eq:approx_krnl_path}), the general functional optimisation given in (\ref{eq:general_object}) is transformed into an optimisation of the weights, 
$\mathbf{w}^\xi_{optimal}= \argmin_{\mathbf{w}^\xi} {U(\xi)}$. The iterative update rule of (\ref{eq:FGMP_update_rule}) takes the following form: 
\begin{equation} \label{eq:update_rule_weights}
	\mathbf{w}^\xi_{n+1} = \mathbf{w}^\xi_n - \eta_n \Lambda^{-1} \hat{\Upsilon}(t_i)^T\hat{\Upsilon}(t_i)\nabla_\xi{U}(\xi_n)(t_i),
\end{equation}
where $t_i \in \mathcal{T}$. 

The algorithm for functional exploration is given in Algorithm \ref{algo:main}. The essential inputs are the initial robot state and the occupancy map. Boundary conditions or an initial solution are optional inputs. In each iteration, a mini-batch is drawn. The safety of each sample is checked, and if it is below $P_{safe}$, the sample is used to update the path. The gradient of the various components (obs, dyn, MI) of the objective functional are computed, and summed according to (\ref{eq:FPMP_U}). Once the overall functional gradient is computed, the weights $\mathbf{w}^\xi_{n+1}$ are updated according to (\ref{eq:update_rule_weights}) .

Figure \ref{fig:iteration} provides an insight into the optimisation process of a single planning iteration. Each iteration starts with an initial guess, which is depicted in blue. The obstacle functional gradient repels the path from obstacles and unknown space. The MI objective pulls the path toward unexplored space. The intermediate path updates, following the functional gradient, are shown in grey. The final optimal path is shown in green.

\begin{figure*}[tpb]
	
	\centering
	
	\includegraphics[width=1.\textwidth]{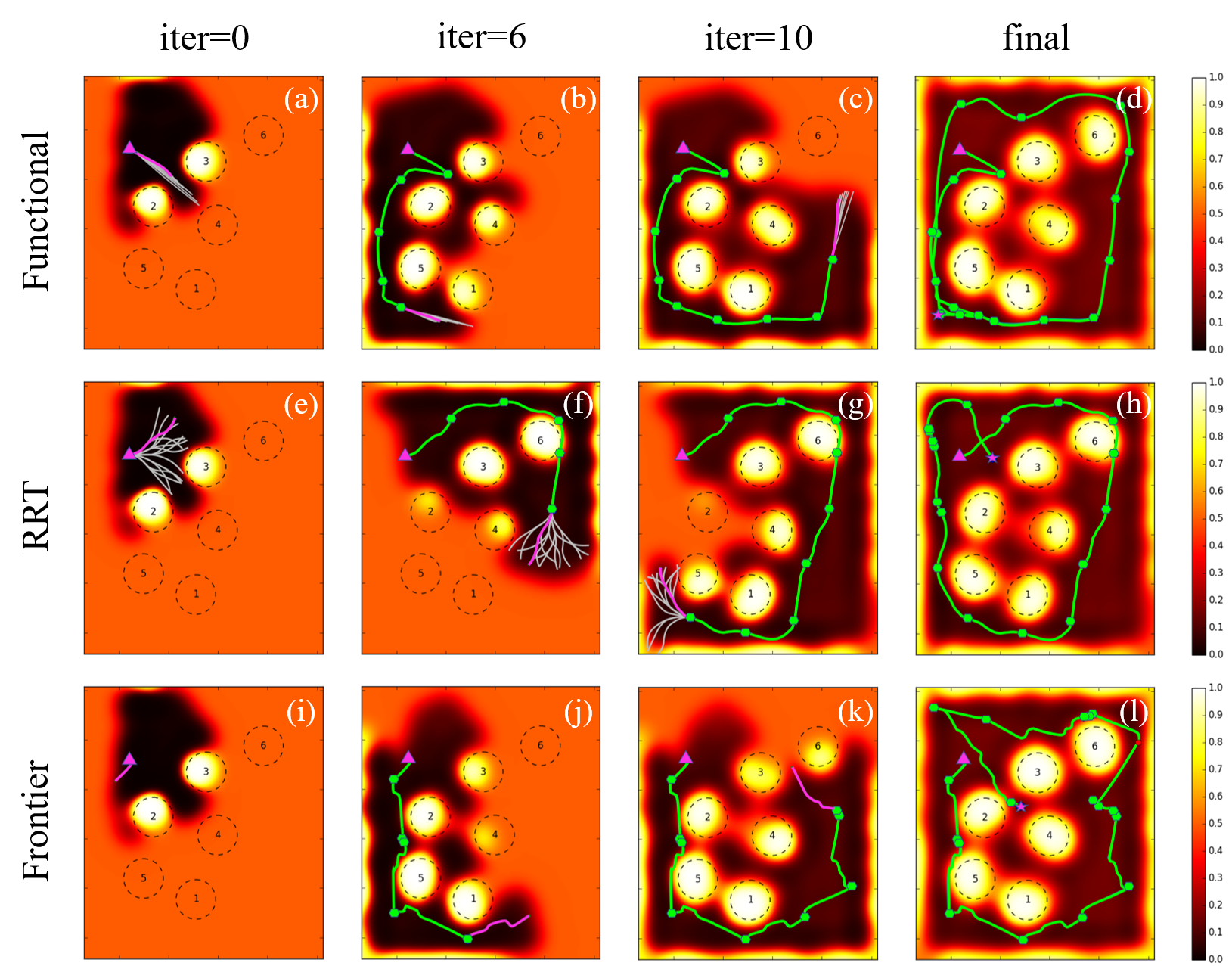}
	
	\caption{Comparison of exploration methods at various planning iterations using continuous occupancy maps. The hexagonal markers are the planning poses. Triangle and star represent start and end points, respectively. The grey lines are either candidate paths (RRT) or intermediate solutions (functional exploration). The pink line depicts the optimised path and the green line is the traversed path. The path is assessed during execution, and a re-plan step is invoked if a path is no longer safe.}
	\label{fig:sim_results}
\end{figure*}

\section{Experimental Results}
\label{sec:results}

In this section, we evaluate the performance of the stochastic functional gradient path planner and compare it to other exploration methods for continuous occupancy maps. As a benchmark, we chose to compare our method to the RRT-based exploration method of \cite{Yang2013} which we modified to use Hilbert maps. This method optimises MI during path selection, and while its bottleneck lies in computational complexity, it takes advantage of the fact the RRTs are probabilistically complete. Another method used for comparison is based on frontier exploration \cite{Yamauchi1997}, which takes a grid approximation of the continuous map in a similar approach to \cite{Jadidi2014}. The path is then constructed by a smooth RRT planner which reasons only about the path safety. All methods, including Hilbert maps, are implemented in Python and tested on an Intel i5-6200U with 8GB RAM.

\subsection{Simulations}
\label{sec:Simulations}

Figure \ref{fig:sim_results} shows a qualitative comparison at various planning iterations. While all exploration methods successfully build the map, there are some clear differences. The frontier-based method  produces jerky paths and tends to move towards the edges and corners of the map. Path optimisation methods, on the other hand, stay closer to obstacles. As the RRT-based method maximises MI only, its path moves closer to the edges of obstacles, while the proposed method keeps a bigger distance, as its objective includes obstacle safety explicitly. We note that increasing the safety margin by applying a blurring filter, as done with grid maps, is not applicable in continuous maps, as it requires expensive discretisation of the map.

\begin{figure}[tpb]
	
	\centering
	
	\includegraphics[width=0.48\textwidth]{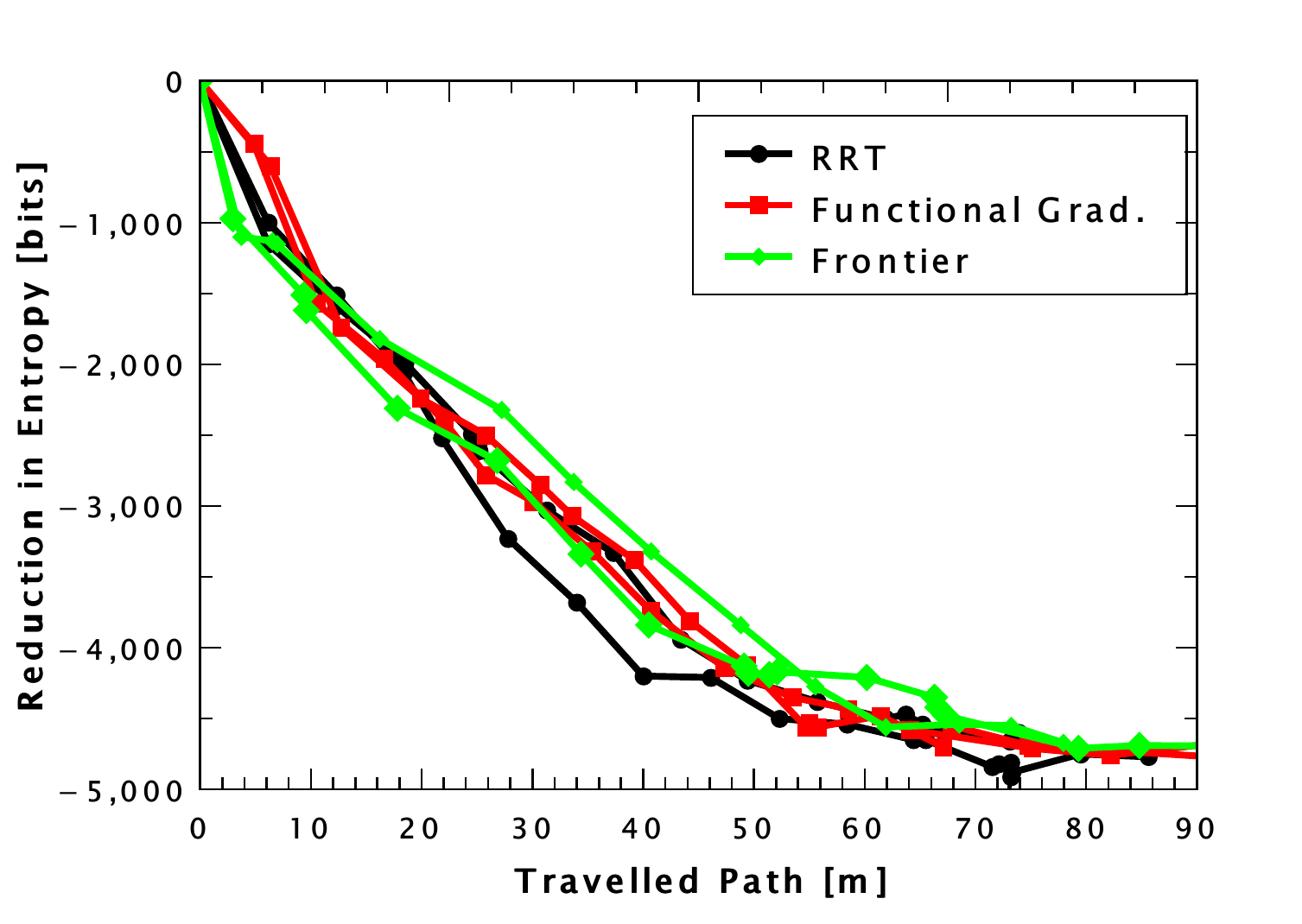}
	
	\caption{Comparison of exploration methods - 2 repetitions; the proposed functional exploration method (red), RRT-based exploration (black) and frontier-based exploration (green). RRT and the propose method converge in a similar fashion as both optimise MI over entire path. The goal-based approach of frontier converges slower as it does not explicitly optimise on MI.}
	\label{fig:comparison}
\end{figure}

Quantitative comparisons are shown in Fig. \ref{fig:comparison} and in Table \ref{tab:performance_comparison}. Fig. \ref{fig:comparison} depicts the reduction in the entropy of the map. The rate of reduction is similar for both our method and the RRT-based exploration, albeit slightly better for the latter, mainly due the fact that the paths generated by the RRT move closer to obstacles. As a result, the RRT planner covers the map faster, however with a higher probability of collision, as shown in Table \ref{tab:performance_comparison} by the maximum occupancy values exceeding the 50\% occupancy threshold. The convergence of the frontier planner is slower, as a result of the over-estimation of the path utility by the choice of a single goal at each iteration. In addition, paths are jerky, leading to longer time to cover the same area. As the frontier planner does not explicitly minimise collision risk, the maximum occupancy over the path is high. In contrast, the maximum occupancy of the proposed method along the path is significantly below the 50\% occupied threshold.

\begin{table}[b]
	\centering
	\caption{Performance comparison (40 planning iterations)}
	\label{tab:performance_comparison}
	\begin{tabular}{lccc}
		\toprule
		& Proposed method &
        RRT \cite{Yang2013} &
         Frontier\cite{Yamauchi1997}  \\
		\midrule
		Mean occupancy & $ 1.2 $ & $ 5.1 $ & $ 3.0 $ \\
        Max. occupancy & $ 26.3 $ & $ 75.2 $ & $ 47.9 $ \\
		Median Iter. Plan Time [s]   & $65.4$ & $200.6$ & $13.4$ \\
   		Mean Iter. Plan Time [s]   & $73.7$ & $196.6$ & $84.3$ \\
        Max. Iter. Plan Time [s]   & $174.4$ & $239.6$ & $1345.2$ \\
        \bottomrule
	\end{tabular}
\end{table}

Comparing the median runtime results shown in Table \ref{tab:performance_comparison} reveals that the RRT planner is significantly slower than our method, mainly because MI is computed over the entire map. Since frontier only queries the map for occupancy, its runtime is significantly smaller. However, its average runtime is similar to our method as occasionally resolving longer path is required. It is worth noting that the runtime results are limited by the Python implementation of the Hilbert map. C++ implementation of the Hilbert map proved to be two to three orders of magnitude faster, which makes it suitable for online applications.     

\subsection{Real World Scenario}
\label{sec:real_world}

\begin{figure}[t]
	
	\centering
	
	\includegraphics[width=0.48\textwidth]{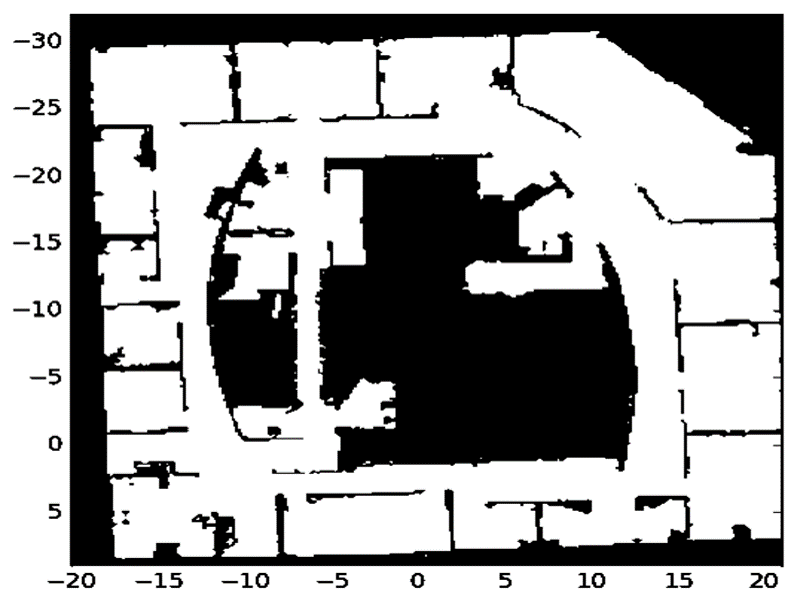}
	
	\caption{Intel-Lab - Ground truth map based on the Intel-Lab dataset. The robot does not have access to this map. It is only used to emulate range observations. }
	\label{fig:intel}
\end{figure}

To evaluate the performance of the functional exploration algorithm in a real world scenario, we simulated a robot exploring the Intel-Lab. We used the Intel-Lab dataset (available at http://radish.sourceforge.net/) to generate a ground truth, shown in Fig. \ref{fig:intel}, from which we can emulate range observations. However, the exploring robot does not have direct access to the ground truth map. The map in Fig. \ref{fig:intel} reveals a relatively simple structure. Yet, the small rooms and narrow corridors pose a difficulty to a robot with limited manoeuvrability. To prevent a situation where the robot is stuck in a room, we added a reverse-on-path option. Meaning, if the robot identifies a dead-end, it may reverse on the path that took it to that spot.
 
\begin{figure*}[tb]
	
	\centering
	\subcaptionbox[width=0.3\linewidth]{1 \label{fig:Intel_explore:subfig-1}}
	{\includegraphics[width=0.3\linewidth]{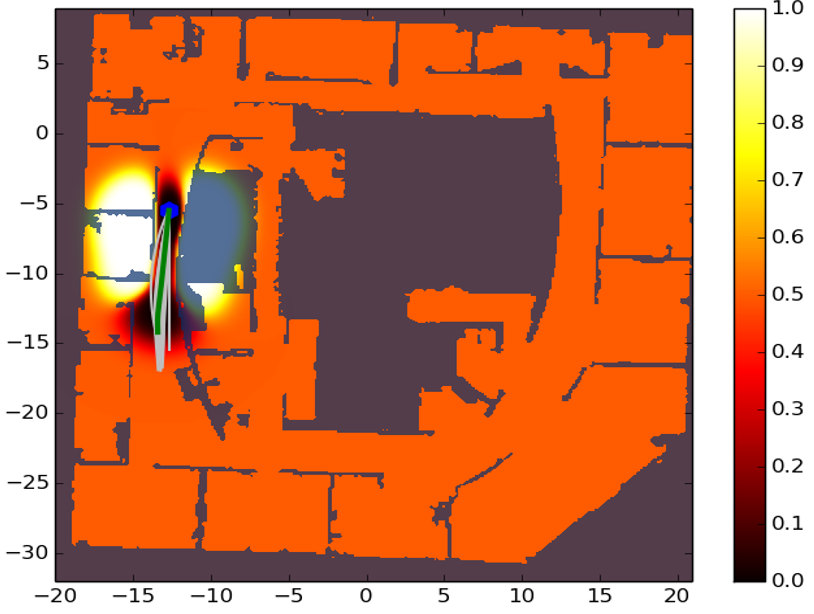}}
	\subcaptionbox[width=0.3\linewidth]{4 \label{fig:Intel_explore:subfig-2}}
	{\includegraphics[width=0.3\linewidth]{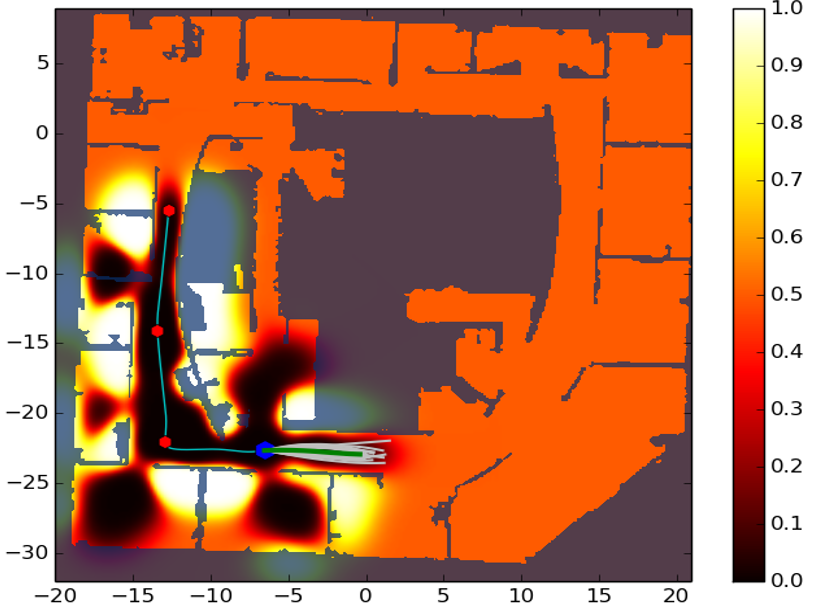}}
	\subcaptionbox[width=0.3\linewidth]{9 \label{fig:Intel_explore:subfig-3}}
	{\includegraphics[width=0.3\linewidth]{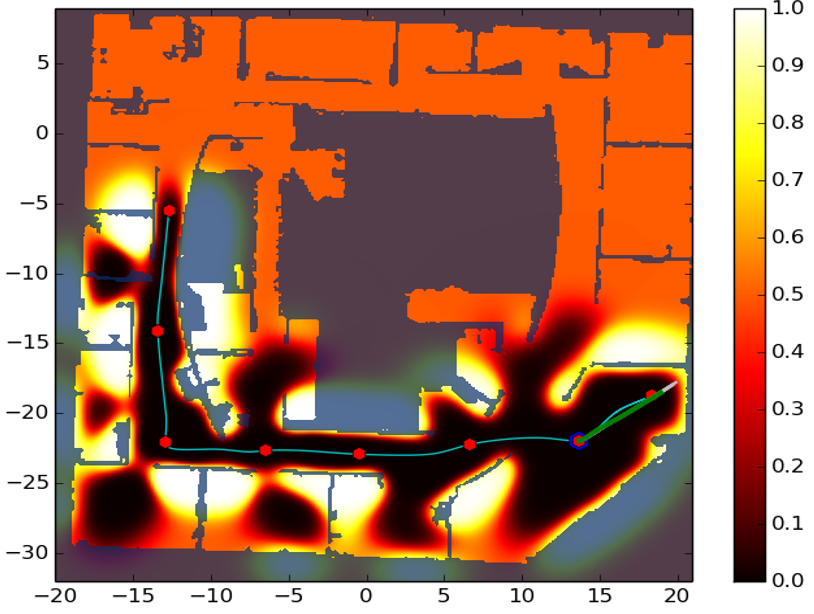}}
	
	\subcaptionbox[width=0.3\linewidth]{14 \label{fig:Intel_explore:subfig-4}}
	{\includegraphics[width=0.3\linewidth]{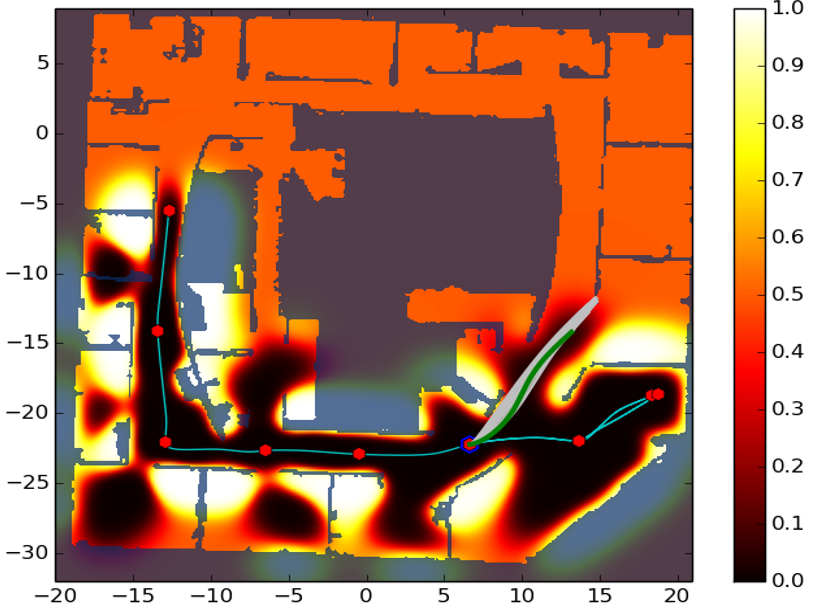}}
	\subcaptionbox[width=0.3\linewidth]{17 \label{fig:Intel_explore:subfig-5}}
	{\includegraphics[width=0.3\linewidth]{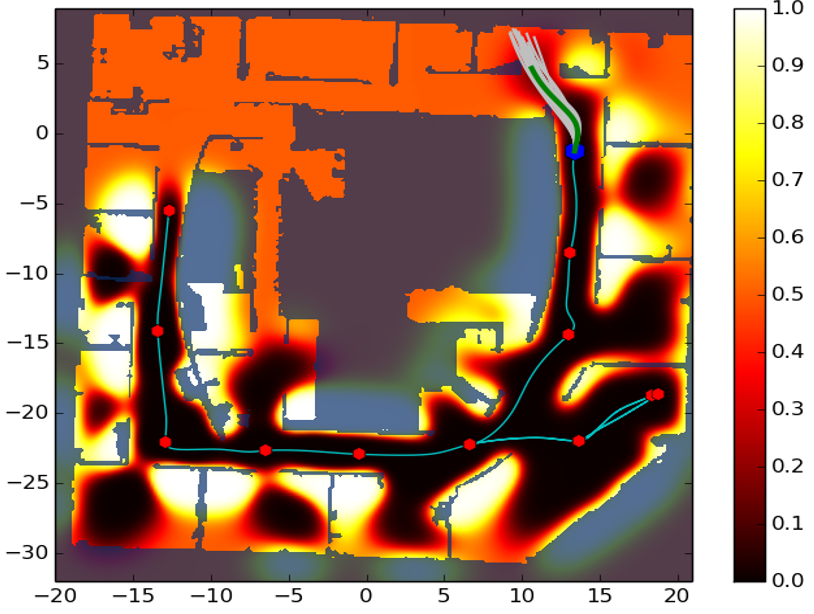}}
	\subcaptionbox[width=0.3\linewidth]{22 \label{fig:Intel_explore:subfig-6}}
	{\includegraphics[width=0.3\linewidth]{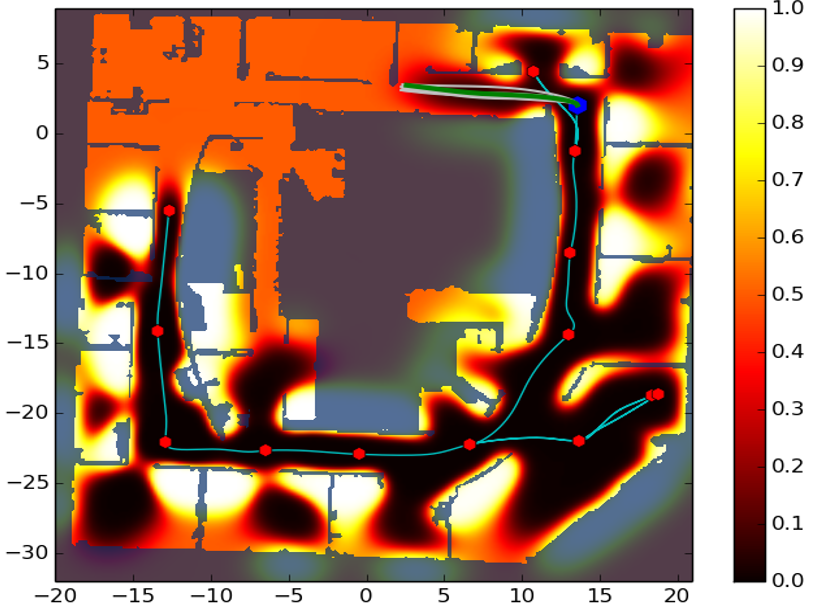}}
	
	\subcaptionbox[width=0.3\linewidth]{25 \label{fig:Intel_explore:subfig-7}}
	{\includegraphics[width=0.3\linewidth]{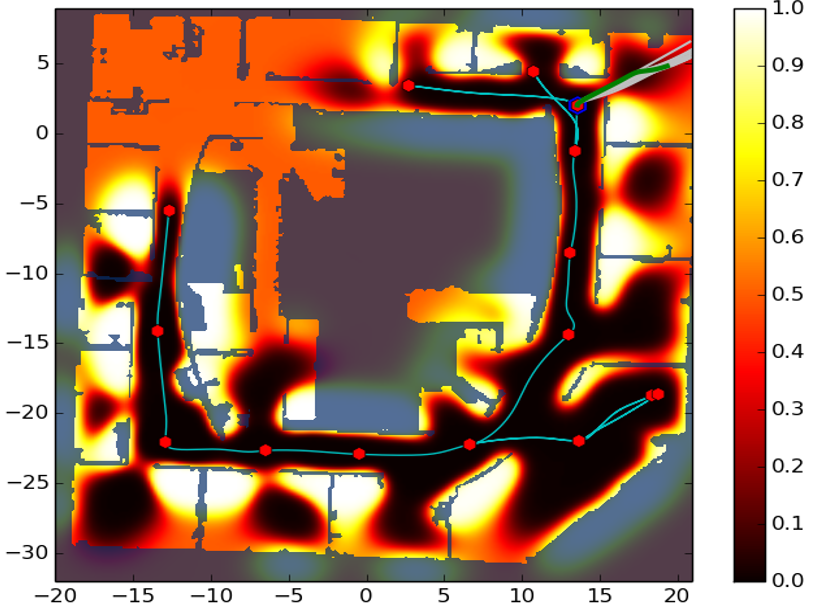}}
	\subcaptionbox[width=0.3\linewidth]{29 \label{fig:Intel_explore:subfig-8}}
	{\includegraphics[width=0.3\linewidth]{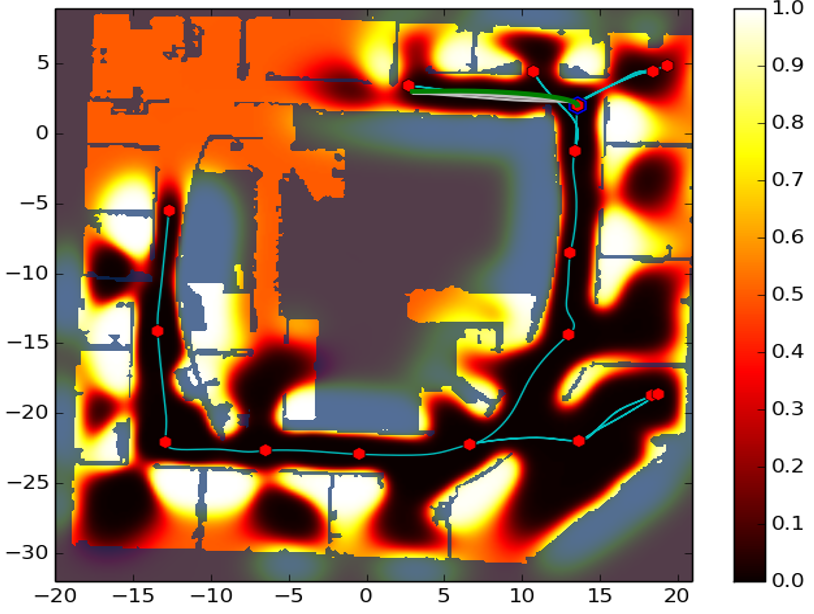}}
	\subcaptionbox[width=0.3\linewidth]{32 \label{fig:Intel_explore:subfig-9}}
	{\includegraphics[width=0.3\linewidth]{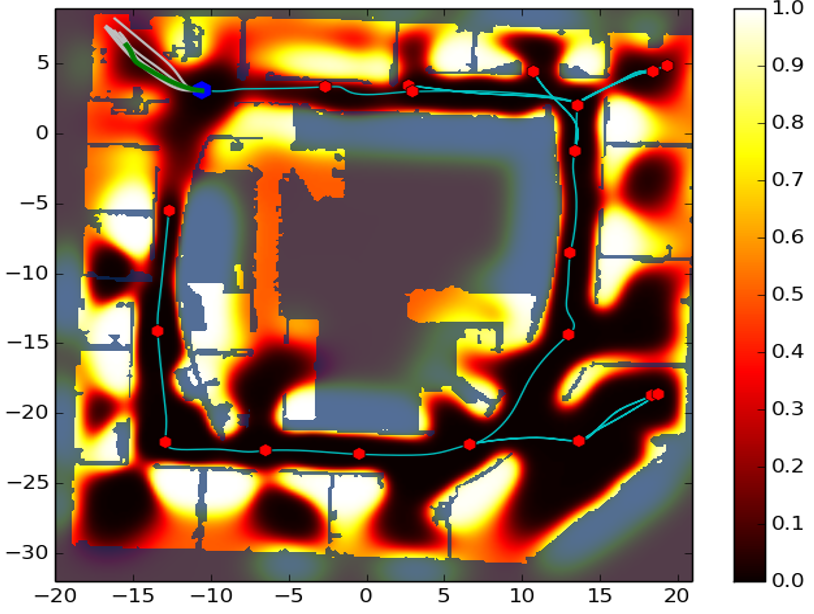}}
	
	\caption{Functional exploration in the Intel-Lab at various planning iterations using continuous occupancy maps.The blue overlay depicts the ground truth, as shown in Fig. \ref{fig:intel}. The grey lines are intermediate solutions and the green line is the optimal path. The path is assessed during execution, and a re-plan step is invoked if a path is no longer safe. The traversed path is plotted in cyan, the red hexagons are the planning poses and the blue hexagon is the current pose. }
	\label{fig:Intel_explore}
\end{figure*}

Figure \ref{fig:Intel_explore} shows the exploration process at various planning iterations. The generated Hilbert map is overlaid with the ground truth map of \ref{fig:intel} as a reference to the map accuracy. The robot successfully explore the majority of the map, moving mainly in the main corridor. It enters only some of the rooms, only where there is enough clearance at the entrance. We note that the robot only relies on occupancy around the entrance to assess safety. 

Figure \ref{fig:Intel_explore} provides an insight to the path optimisation process. This is shown by the intermediate paths (in grey), which reveals how the functional objective in Eq. (\ref{eq:FPMP_U}) balances safety with exploration during the path selection process. The MI term in Eq. (\ref{eq:FPMP_U}) pulls paths towards the border between known and unknown space. The safety functional, on the other hand, maintains a safe distance from obstacles and unknown space. Consequently, paths tend to move in the middle of the corridors and end close, but within some margin, to a frontier. 

The main limitation of the functional exploration approach is the lack of global context during the optimisation. As FGD in a local optimisation process, its outcome depends on the starting point of the optimisation. This make FGD sensitive to dead-ends. An exploration dead-end scenario is shown in Fig. \ref{fig:Intel_explore:subfig-3}, where a robot is inside a room unable to find its way out. When the robot is inside a room, it can not identify any planning horizon in its local neighbourhood. Meaning, the MI functional's contributions during optimisation are negligible, which results in non-exploring paths. To somewhat resolve this problem, we added a reverse-on-path option when the algorithm identifies a dead-end. However, a more robust solution may include a global exploration initial guess, such as a frontier, to start the optimisation. This will require to develop a frontier detection method for continuous occupancy maps, as current frontier exploration methods require to discretise the occupancy map, which is computationally intensive.

\section{Conclusions}
\label{sec:conclusions}

This paper introduces a novel method for exploration over continuous occupancy maps using stochastic functional gradient descent. This approach formalises exploration as a variational problem, where optimisation is performed directly in the space of trajectories. The functional objective of the proposed method explicitly optimises both safety and information collection over the entire path, finding the \textit{Next Best Path}. While this approach can be used with any type of occupancy map, it is highly effective with Hilbert maps, where the introduced MI objective and its gradient can be computed from a perturbed model of the map. Our proposed approach eliminates the need for computing MI over the entire map as done in other exploration techniques. Rather, it computes variations to the path based on  functional gradient of MI which are efficiently derived in closed-form from the map model.

Comparisons with other exploration methods show that the proposed method improves on both safety and MI. Point exploration methods, such as frontier, which do not optimise the path selection, exhibit slower exploration rate. On the other hand, sampling-based exploration methods, such as \cite{Yang2013}, do not include safety in their objective, hence the resulting paths tends to move closer to obstacles. Moreover, these methods are computationally expensive due to the need to repeatedly sample the MI objective over the entire path. In comparison our proposed method achieves similar exploration rates to \cite{Yang2013} while taking less time to compute and  still maximising safety.

\bibliographystyle{ieeetr}
\bibliography{ISRR2017c}

\begin{thebibliography}{10}

\bibitem{Stachniss2009}
C.~Stachniss, {\em {Robotic Mapping and Exploration}}.
\newblock Springer, 2009.

\bibitem{Thrun2005a}
S.~Thrun, W.~Burgard, and D.~Fox, {\em {Probabilistic Robotics}}.
\newblock MIT Press, 2005.

\bibitem{Julia2012}
M.~Juli{\'{a}}, A.~Gil, and O.~Reinoso, ``{A Comparison of Path Planning
  Strategies for Autonomous Exploration and Mapping of Unknown Environments},''
  {\em Autonomous Robots}, vol.~33, no.~4, pp.~427--444, 2012.

\bibitem{Yamauchi1997}
B.~Yamauchi, ``{A Frontier-based Approach for Autonomous Exploration},'' in
  {\em Proceedings of the IEEE International Symposium on Computational
  Intelligence in Robotics and Automation}, 1997.

\bibitem{Gonzalez-Banos2002}
H.~H. Gonz{\'{a}}lez-Ba{\~{n}}os and J.-C. Latombe, ``{Navigation Strategies
  for Exploring Indoor Environments},'' {\em The International Journal of
  Robotics Research}, vol.~21, no.~10-11, pp.~829--848, 2002.

\bibitem{Holz2011}
D.~Holz, N.~Basilico, F.~Amigoni, and S.~Behnke, ``{A Comparative Evaluation of
  Exploration Strategies and Heuristics to Improve Them},'' in {\em Proceedings
  of the European Conference on Mobile Robots}, 2011.

\bibitem{Elfes1996}
A.~Elfes, ``{Robot Navigation: Integrating Perception, Environmental
  Constraints and Task Execution within a Probabilistic Framework},'' in {\em
  Reasoning with Uncertainty in Robotics}, 1996.

\bibitem{Whaite1997}
P.~Whaite and F.~P. Ferrie, ``{Autonomous Exploration: Driven by
  Uncertainty},'' {\em IEEE Transactions on Pattern Analysis and Machine
  Intelligence}, vol.~19, no.~3, pp.~193--205, 1997.

\bibitem{Vallve2015}
J.~Vallv{\'{e}} and J.~Andrade-Cetto, ``{Potential Information Fields for
  Mobile Robot Exploration},'' {\em Robotics and Autonomous Systems}, vol.~69,
  pp.~68--79, 2015.

\bibitem{Charrow-RSS-15}
B.~Charrow, G.~Kahn, S.~Patil, S.~Liu, K.~Goldberg, P.~Abbeel, N.~Michael, and
  V.~Kumar, ``{Information-Theoretic Planning with Trajectory Optimization for
  Dense 3D Mapping},'' in {\em Proceeding of Robotics: Science and Systems},
  2015.

\bibitem{Lauri2015}
M.~Lauri and R.~Ritala, ``{Planning for Robotic Exploration based on Forward
  Wimulation},'' {\em preprint arXiv:1502.02474}, 2015.

\bibitem{Yang2013}
K.~Yang, S.~{Keat Gan}, and S.~Sukkarieh, ``{A Gaussian Process-based RRT
  Planner for the Exploration of an Unknown and Cluttered Environment with a
  UAV},'' {\em Advanced Robotics}, vol.~27, no.~6, pp.~431--443, 2013.

\bibitem{Jadidi2014}
M.~G. Jadidi, J.~V. Miro, R.~Valencia, and J.~Andrade-Cetto, ``{Exploration on
  Continuous Gaussian Process Frontier Maps},'' in {\em Proceedings of the IEEE
  International Conference on Robotics and Automation}, 2014.

\bibitem{Jadidi2015}
M.~G. Jadidi, J.~V. Miro, and G.~Dissanayake, ``{Mutual Information-based
  Exploration on Continuous Occupancy Maps},'' in {\em Proceedings of the
  IEEE/RSJ International Conference on Intelligent Robots and Systems}, 2015.

\bibitem{Marchant2014}
R.~Marchant and F.~Ramos, ``{Bayesian Optimisation for Informative Continuous
  Path Planning},'' in {\em Proceedings of the IEEE International Conference on
  Robotics and Automation}, 2014.

\bibitem{Francis2017a}
G.~Francis, L.~Ott, R.~Marchant, and F.~Ramos, ``{Occupancy Map Building
  through Bayesian Exploration},'' {\em preprint arXiv:1703.00227}, 2017.

\bibitem{Francis2017}
G.~Francis, L.~Ott, and F.~Ramos, ``{Stochastic Functional Gradient for Motion
  Planning in Continuous Occupancy Maps},'' in {\em Proceeding of the IEEE
  International Conference on Robotics and Automation}, 2017.

\bibitem{Francis2017b}
G.~Francis, L.~Ott, and F.~Ramos, ``{Stochastic Functional Gradient Path
  Planning in Occupancy Maps},'' {\em preprint arXiv:1705.05987}, may 2017.

\bibitem{ramos2015hilbert}
F.~Ramos and L.~Ott, ``{Hilbert maps: Scalable Continuous Occupancy Mapping
  with Stochastic Gradient Descent},'' in {\em Proceedings of Robotics: Science
  and Systems}, 2015.

\bibitem{Elfes1989}
A.~Elfes, ``{Using Occupancy Grids for Mobile Robot Perception and
  Navigation},'' {\em Computer}, vol.~22, no.~6, pp.~46--57, 1989.

\bibitem{OCallaghan2012}
S.~T. O'Callaghan and F.~T. Ramos, ``{Gaussian Process Occupancy Maps},'' {\em
  The International Journal of Robotics Research}, vol.~31, no.~1, pp.~42--62,
  2012.

\bibitem{Zucker2013}
M.~Zucker, N.~Ratliff, A.~D. Dragan, M.~Pivtoraiko, M.~Klingensmith, C.~M.
  Dellin, J.~A. Bagnell, and S.~S. Srinivasa, ``{CHOMP: Covariant Hamiltonian
  Optimization for Motion Planning},'' {\em The International Journal of
  Robotics Research}, vol.~32, no.~9-10, pp.~1164--1193, 2013.

\bibitem{Marinho2016}
Z.~Marinho, B.~Boots, A.~Dragan, A.~Byravan, G.~J. Gordon, and S.~Srinivasa,
  ``{Functional Gradient Motion Planning in Reproducing Kernel Hilbert
  Spaces},'' in {\em Proceedings of Robotics: Science and Systems}, 2016.

\bibitem{Julian2014}
B.~J. Julian, S.~Karaman, and D.~Rus, ``{On Mutual Information-based Control of
  Range Sensing Robots for Mapping Applications},'' {\em The International
  Journal of Robotics Research}, vol.~33, no.~10, pp.~1375--1392, 2014.

\bibitem{mukadam2016gaussian}
M.~Mukadam, X.~Yan, and B.~Boots, ``{Gaussian Process Motion Planning},'' in
  {\em Proceedings of the IEEE International Conference on Robotics and
  Automation}, 2016.

\bibitem{Dong2016}
J.~Dong, M.~Mukadam, F.~Dellaert, and B.~Boots, ``{Motion Planning as
  Probabilistic Inference using Gaussian Processes and Factor Graphs},'' in
  {\em Proceedeing of Robotics: Science and Systems}, 2016.

\bibitem{Robbins1951}
H.~Robbins and S.~Monro, ``{A Stochastic Approximation Method},'' {\em The
  Annals of Mathematical Statistics}, vol.~22, no.~3, pp.~400--407, 1951.

\end{thebibliography}

\end{document}